\documentclass[11pt,a4paper]{article}
\usepackage{acl2023}
\usepackage{times}
\usepackage{latexsym}
\usepackage{tabularx}
\usepackage{framed}
\usepackage{lipsum}
\usepackage{mathrsfs}
\usepackage{amsmath}
\usepackage{arydshln}
\usepackage{pgfplots}
\usepackage{tikz}
\pgfplotsset{
   compat=1.17,
   legend entry/.initial=,
   every axis plot post/.code={%
       \pgfkeysgetvalue{/pgfplots/legend entry}\tempValue
       \ifx\tempValue\empty
           \pgfkeysalso{/pgfplots/forget plot}%
       \else
           \expandafter\addlegendentry\expandafter{\tempValue}%
       \fi
   },
}
\usepackage[ruled]{algorithm2e}

\usepackage{tikz}
\usepackage{circledsteps}
\newsavebox{\tempbox}
\newcommand{\textbox}[1]
{\savebox{\tempbox}{#1}
 \ifdim\wd\tempbox<\columnwidth\relax
   \makebox[1.2\columnwidth]{\usebox{\tempbox}}%
 \else
   \parbox{1.1\columnwidth}{\raggedright #1}%
 \fi}
\usepackage{microtype}
\usetikzlibrary{decorations.pathreplacing}

\title{Investigating a Benchmark for Training-set free Evaluation of\\Linguistic Capabilities in Machine Reading Comprehension}

\author{Viktor Schlegel, {Goran Nenadic} \and {Riza Batista-Navarro} \\
Department of Computer Science, The University of Manchester, UK 
\\
\texttt{\{viktor.schlegel, riza.batista, g.nenadic\}@manchester.ac.uk
}}

\date{}

\begin{document}
\maketitle
\begin{abstract}
Performance of NLP systems is typically evaluated by collecting a large-scale dataset by means of crowd-sourcing to train a data-driven model and evaluate it on a held-out portion of the data. This approach has been shown to suffer from spurious correlations and the lack of challenging examples that represent the diversity of natural language. 
Instead, we examine a framework for evaluating optimised models in training-set free setting on synthetically generated challenge sets. 
We find that despite the simplicity of the generation method, the data can compete with crowd-sourced datasets with regard to naturalness and lexical diversity for the purpose of evaluating the linguistic capabilities of MRC models. We conduct further experiments and show that state-of-the-art language model-based MRC systems can learn to succeed on the challenge set correctly, although, without capturing the general notion of the evaluated phenomenon. 
\end{abstract}

\section{Introduction}

Research on Machine Reading Comprehension (MRC), the task of finding an answer to a question referring to a passage of text, has gained noticeable traction in recent years, with recent approaches surpassing human performance on multiple benchmarks \cite{Lan2020,Raffel2019}. Due to its flexibility, MRC is a popular formulation to establish various natural language understanding (NLU) capabilities, such as multi-hop \cite{Yang2018}, logical \cite{Liu2020c} and arithmetic \cite{Dua2019} reasoning, among others. MRC benchmarks are designed such that a model is fine-tuned on a large, task-specific dataset featuring a phenomenon of interest (such as those mentioned before) and evaluated on a held-out portion of that data. If the model achieves good performance on that evaluation set, it is often concluded that the model learned to process this phenomenon correctly.

This in-distribution approach to evaluating NLU is problematic for different reasons \cite{Linzen2020} for example, it clearly favours data-driven approaches that are potentially pre-trained on arbitrary amounts of data, such as contemporary transformer-based language models \cite{Devlin2018}. 
To scale the size of these benchmarks to allow data-driven approaches to be optimised upon them, these datasets are collected by means of crowd-sourcing. The lack of linguistic expertise amongst lay annotators leads to the absence of challenging phenomena in MRC data \cite{Sugawara2017,Sugawara2019,Schlegel2020}. Moreover, simplifying heuristics adapted by annotators result in ``annotations artefacts'' \cite{gururangan2018annotation}, i.e., cues in data that are spuriously predictive of the expected answer. It has been shown that low-bias, data-driven approaches such as neural networks follow the strongest signal present in training data, and learn to rely on the presence of such artefacts \cite{Jia2017,Sugawara2018,Ko2020}. 
This reliance is not exposed following the conventional in-distribution evaluation, because by design, these cues appear in both training and evaluation data. Thus, a model can achieve high performance by relying on their presence, potentially circumventing the requirement to perform natural language understanding of the evaluated phenomenon \cite{Schlegel2020a}.



\begin{figure}[t]
    \centering
    \begin{tabularx}{0.98\columnwidth}{rX}
    \hline
    \footnotesize {Orig.:} & \footnotesize{ She scored a goal from 25 metres away.} \\
    \footnotesize {SAM:} & \footnotesize{ She {\color{blue}\emph{almost}} scored a goal from 25 metres away.} \\
    \footnotesize {SPM:} & \footnotesize {She scored a goal from \emph{\color{orange} almost} 25 metres away.} \\
    \hline
    \hline
    \multicolumn{2}{c}{\footnotesize{\emph{Can MRC models determine whether a goal was scored?}}} \\
    \hline
\end{tabularx}
    \caption{Example of an {\color{blue}altering} (SAM) and  {\color{orange}preserving} (SPM) modification of a sentence probing whether MRC models learn to process their semantics correctly.}
    \label{fig:intro}
\end{figure}

To alleviate this issue, optimised models can be evaluated on so called \emph{challenge sets} that feature a phenomenon of interest.  Because they stem from a different generative process, 
cues present in training data are unlikely to appear in the challenge set as well, and reliance on their exploitation does not lead to good performance. Challenge sets, however, confound two sources of error, namely errors due to the incapability of a model to process the investigated phenomenon and errors due to the distribution shift between training and challenge set data.
For example, an MRC model optimised on trivia questions might have never encountered a question regarding football match reports, and is more likely to err due to diverging vocabulary and different discourse structure. 

Recently, a domain-independent evaluation methodology was proposed, in an attempt to distinguish between these sources of error \cite{Schlegel2021}. The authors evaluate the capabilities of optimised MRC models to distinguish between lexically similar yet semantically different sentences independent of their training data, using synthetic challenge set data. 
In the present paper, we conduct experiments as an in-depth analysis of their challenge set generation method and their resulting data. We provide empirical evidence that models can learn to succeed on the challenge set from relatively few examples, without loss of in-distribution performance of optimised models. We show, however, that this success does not represent the capability to process the underlying phenomenon but is rather due to quick overfitting to the patterns inherent to the challenge set data. 

\section{Background: Challenge set and Metric}
This section briefly summarises the introduced challenge set and evaluation metric. For more details, consult \citet{Schlegel2021}.

Intuitively, each example in a contrastive challenge set consists of two versions---a \emph{Baseline} version, and an \emph{Intervention} version with a small but meaningful modification such that it changes the expected prediction.  
For the MRC task this means that their presence changes the expected answer. 
More formally, for each MRC challenge set baseline instance $B_i=(Q_i,P_i,A_i)$ consisting of question $Q_i$, passage $P_i$ and expected answer $A_i$, there exists a corresponding \emph{aligned} intervention instance $I_i=(Q_i,P'_i,A'_i)$ that differs from the baseline instance only in that $P'_i$ contains a modification and $A_i\neq A'_i$, because the modification changes the expected answer. A contrastive challenge set with $N$ baseline and intervention is described as  $\mathcal{N} = (\mathcal{B}, \mathcal{I})$.

This formulation allows to evaluate the capability to correctly process examples in both baseline and modified form, referred to as $|\mathcal{B}^+\cap \mathcal{I}^+|$. Thus \emph{consistency} is defined as \mbox{$C(f_{\theta}) = \frac{|\mathcal{B}^+\cap \mathcal{I}^+|}{|\mathcal{N}|}$}, of an evaluated model $f_{\theta}$, with challenge set size $|\mathcal{N}|$. A more lenient version of consistency is $
DICE(f_{\theta}) = \frac{|\mathcal{B}^+\cap \mathcal{I}^+|}{|\mathcal{B}^+|}
$.
This measure only takes into account those examples that were correctly processed in their \emph{baseline} form. This establishes the baseline performance and discounts for all errors that an optimised model makes based on the differences between the distribution of training and challenge set data and highlights errors made due to the presence of the modification. The underlying, implicit assumption is that these errors occur independently of each other. We provide empirical evidence for this assumption. 


In the investigated challenge set, intervention instances contain \emph{Semantics Altering Modifications} (SAM)---linguistic phenomena that change the semantics of a sentence when inserted, as exemplified in the first two rows of Figure~\ref{fig:intro}. 
These are fairly regular and closed-class expressions and their semantics can be consistently described as follows in lambda calculus: \mbox{$sam = \lambda v. \neg v$}. That is, $sam$ appears before a verb phrase $\lambda z. VP(z)$ and negates its meaning, s.t. the resulting expression evaluates to $\lambda z. \neg VP(z)$.

\section{Experiments}

In this section, we reproduce the dataset used by \newcite{Schlegel2021} and evaluate the quality of the generated passages by comparison to real-world passages. 
Furthermore, we investigate whether ``inoculating'' \cite{Liu2019b} models on portions of the generated data can help it to succeed at the challenge set and whether doing so reliably reflects the capability of processing SAM.

\textbf{Data is unexpectedly diverse.}
We compare the generated paragraphs with the topically most similar existing MRC data: a subset of the DROP dataset \cite{Dua2019} that features questions about NFL match reports and similar question types.
Specifically, we compare them in terms of \emph{Diversity}, i.e. average of the Jaccard similarity between each two pairs of passages, to see how (lexically) different the generated passages are, and \emph{Naturalness} to estimate how realistic they are. 
For \emph{Naturalness}, we calculate the average sentence-level cohesion indices that were reported to correlate with human judgements of essay writing quality \cite{Crossley2016,Crossley2019}. 
We establish the measures using the associated \texttt{TAACO} tool 
and report results in Table~\ref{tab:naturality}.

\begin{table}[t]
\centering
    \begin{tabularx}{0.98\columnwidth}{X c c}

\textbf{Measure} & \textbf{SAM} & \textbf{NFL} \\
\hline 
\footnotesize{\emph{higher is better} $\uparrow$} & & \\
$m_1$ Adj. sentence similarity & $0.58$ & $0.67$ \\
\hline
\hline
\footnotesize{\emph{lower is better} $\downarrow$}  & & \\
$m_2$ Type-token ratio & $0.72$ & $0.66$ \\
$m_3$ Adj. sentence verb overlap  & $0.17$ & $0.24$ \\
$m_4$ Pronoun-noun-ratio & $0.07$ & $0.05$ \\
\hdashline
Lexical Diversity & $0.22$ & $0.16$ \\
\hline

\end{tabularx}
\caption{Indices that evaluate the \emph{Naturalness} and \emph{Diversity} of SAM data and the DROP-NFL dataset.} 
\label{tab:naturality}
\end{table}
The low difference between synthetic and natural data is surprising, taking into account the low number of templates and the small and simple generative grammar. For example, the verb overlap between adjacent sentences is lower in the synthetic data, despite the fact that all action verbs stem from a fixed-size vocabulary. The lower measures of semantic similarity of adjacent sentences, the type-token-ratio and the lexical diversity in natural data, hint at the broader coverage of the NFL human-written passages, often including sentences such as \emph{``Coming off their win over the Eagles, the Patriots returned home for an AFC duel with the Indianapolis Colts.''}, which do not directly describe events that occurred during the matches. 

\textbf{Models obtain good performance from few SAM examples.}
To investigate the factors that lead to success on the challenge set, we evaluate a \texttt{roberta-base} \cite{Liu2019c} MRC model that was optimised on different combinations of training data. 
Unless stated otherwise, ``seed'' templates were distinctly split for generating augmentation and challenge set data. All results presented in the remainder of the paper were obtained by averaging the scores of three models trained with different random seeds, each of them evaluated on five versions of the SAM challenge sets featuring $4200$ aligned instances, generated from different random seeds. For the binary scores we report the error margin as a confidence interval at $\alpha=0.05$ using asymptotic normal approximation. Low intervals indicate that obtained scores are in fact stable and not due to chance.

\begin{table}[!tb]
\centering
    \begin{tabular}{l c c c}
\textbf{Training data} & \textbf{EM} & \textbf{F1} & $DICE$ \\
\hline
\textsc{SQuAD} & $82.70$ & $90.36$  & $11 \pm 1$ \\
\textsc{SQuAD}$+ \mathcal{B}$ & $83.27$ & $90.62$ & $9 \pm 2$ \\
\textsc{SQuAD}$+\mathcal{I}$& $83.18$ &$90.58$ &  $92 \pm 1$ \\
\textsc{SQuAD}$+\mathcal{B} + \mathcal{I}$& $83.17$ &$90.60$ &  $87 \pm 1$ \\
\hline
\end{tabular}
\caption{Challenge set $DICE$ score, \textsc{SQuAD} development set accuracy (EM) and (token-level) F1 scores of \texttt{roberta-base} optimised on \textsc{SQuAD} with $1000$ baseline ($\mathcal{+B}$) or (and) intervention ($\mathcal{+I}$) examples. 
}
\label{tab:res-aug}
\end{table}

First, we optimise models on the SQuAD \cite{rajpurkar2016squad} dataset randomly augmented \cite{Liu2019b} with $1000$ baseline examples ($+ \mathcal{B}$), $1000$ intervention examples ($+ \mathcal{I}$) and 500 examples from both sets ($+\mathcal{B}+\mathcal{I}$) and measure the $DICE$ score.
The results are reported in Table~\ref{tab:res-aug}. 
The models optimised on the dataset that was augmented both with baseline and intervention examples learn to answer instances of both types correctly, as evidenced by a high $DICE$ score of $92 \pm 1$, a significant improvement over the optimised model that did not encounter any challenge set examples during training ($DICE$ score of $11 \pm 1$).
Intuitively, the low result of the model trained on \textsc{SQuAD}$+ \mathcal{B}$ is expected and similar to what \newcite{Kaushik2019a} report: as the model only encounters non-altered in-domain examples, it is not incentivised to learn to process SAM correctly. 
This contributes empirical evidence to the assumption from Section~2: errors due to processing SAM incorrectly and errors due to distribution shift are (approximately) independent for the SAM challenge set. In other words, the accuracy on the baseline version of the challenge set represents the ability of an optimised model to transfer to the challenge set data and, for those examples, where the model succeeds, the difference between performance on baseline and intervention instances represents the capability to process SAM. 

\begin{figure}[!tb]
\begin{center}
\begin{tikzpicture}
\begin{axis}[
   title={},
   width=1.1\columnwidth,
   height=0.53\columnwidth,
   legend pos=south east,
   legend columns=3, 
   ymajorgrids=true,
   grid style=dashed,
]
\addplot [
    color=blue,
    mark=square,
    dashed,
    legend entry=qtypes
]
    table [
    x=Step, 
    y=roberta-base-squad-intervention-aug-split-reasoning - eval/dice/, 
    col sep=comma, 
    ignore chars={"},
    ]
  {experiment.csv};
\addplot [
    color=blue,
    mark=o,
    legend entry=qtypes-2
]table [
    x=Step, 
    y=roberta-base-squad-intervention-aug-split-reasoning-2 - eval/dice/,
    col sep=comma, 
    ignore chars={"},
] {experiment.csv};
\addplot [   
    color=orange,
    mark=square,
    dashed,
    legend entry=sam
]
table [
    x=Step, 
    y=roberta-base-squad-intervention-aug-split-sam - eval/dice/, 
    col sep=comma, 
    ignore chars={"},
] {experiment.csv};
\addplot [
    color=orange,
    mark=square,
    legend entry=sam-2
] table [
    x=Step, 
    y=roberta-base-squad-intervention-aug-split-sam-hard - eval/dice/, 
    col sep=comma, 
    ignore chars={"},
] {experiment.csv};
\addplot [
    color=green,
    mark=x,
    legend entry=\#-sam
]table [
    x=Step, 
    y=roberta-base-squad-intervention-aug-split-num-sam - eval/dice/, 
    col sep=comma, 
    ignore chars={"},
] {experiment.csv};

\end{axis}
\end{tikzpicture}
\end{center}
\caption{$DICE$ score on the challenge set as a function of the number of intervention examples mixed into the \textsc{SQuAD} training set.}
\label{fig:data-aug}
\end{figure}
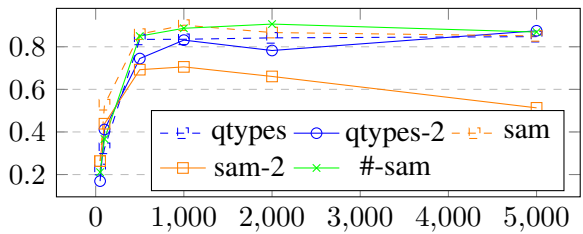

We take the previous idea further and investigate \emph{how many} training SAM examples are needed to perform well on the challenge set. To that end, we train MRC models on the \textsc{SQuAD} dataset, augmented with $50$, $100$, $500$, $1000$, $2000$ and $5000$ 
intervention examples. When generating augmentation and evaluation data, 
we split randomly by question types (\emph{qtypes} and \emph{qtypes-2} in Figure~\ref{fig:data-aug}), SAM category (\emph{sam} and \emph{sam-2})
as well as number of SAM encountered in the passage -- one for augmentation examples, two and three for evaluation data (\emph{\#-sam}). The results in Figure~\ref{fig:data-aug} indicate that between 500 and 1000 examples are enough to perform reasonably well ($DICE$ score of $> 0.8$), even on question types, number and type of SAM not encountered during training. The only exception to this appears to be when adverbial modifiers (e.g. \emph{``almost''} or \emph{``nearly''}) are not encountered during training (\emph{sam-2} split): the performance deteriorates as more examples of other SAM categories are used to augment the training set. 

\textbf{Inoculated models do not learn SAM.} Finally, we investigate whether success of ``inoculated'' models on the challenge set data is truly representative of the capability to process SAM correctly. For simplicity, we confine our investigation to one category of SAM---adverbial modifications with words like ``almost''. We construct a separate evaluation set by changing the position of modifiers such that they do not modify the expected answer, i.e. preserve the semantics of the modified sentence (c.f. Figure~\ref{fig:intro}, third sentence). Intuitively, a model that learned to process the 
SAM correctly, should be able to distinguish whether their presence in fact modifies the semantics of the sentence and thus the expected answer. Therefore, it should retrieve correct answers for sentences with both semantics altering and preserving modifications (SPM).

To investigate this, we evaluate models optimised on SQuAD augmented with 500 $\mathcal{B}$ and 500 $\mathcal{I}$ instances and in addition to the $DICE$ score, we measure their performance on the SPM challenge set. While the models achieve high $DICE$ score of $94\pm1$ and solve $64\%$ and $75\%$ of $\mathcal{B}$ and $\mathcal{I}$ instances, respectively, they only solve $0.03\%$ of the SPM examples. This indicates, that instead of learning the semantics of the words from these categories, 
they quickly learn to disregard sentences containing these words as possible answer candidates alltogether.

\section{Discussion and Conclusion}
In this paper we address two recently proposed suggestions to evaluate MRC models: firstly, the use of expert-curated benchmarks that make explicit the requirements associated with them by design and, secondly, training set-free evaluation \cite{Sugawara2021BenchmarkingPerspective,Linzen2020}.
To that end, we investigate a recently proposed challenge set methodology and discuss metrics that can be used to establish the quality of the automatically generated data. 
We conduct experiments which suggest requirements for MRC training data, such that neural MRC models, that are optimised on these data, learn to succeed at the SAM challenge set. 

A problem associated with controlled experiments such as presented here is that they bear negative predictive power only: 
we can only use challenge sets to identify gaps in the capabilities of the state-of-the-art rather than take high scores as evidence for the acquisition of said capabilities. This is the case for many other proposed evaluation methodologies as well, such as partial input baselines \cite{Feng2019} and in a broader sense with all empirical research. Furthermore, similar to \newcite{Rozen2019} and 
\cite{Schlegel2022CanLanguage}, we caution to interpret good performance of models ``inoculated'' on challenge set as evidence that the models in fact learned to process that phenomenon: it is more likely that the model quickly learns to exploit the patterns of the (often regular) challenge set. Thus, \textbf{challenge sets should be used without inoculation}, and performance improvements should be derived from either architectural improvements or augmentation data drawn from other sources.




The methodology can be generalised to generate challenge sets for different phenomena, as long as passages can be modified to contain the features. Examples of other phenomena that are not well represented in MRC evaluation data thus far are \emph{dative} and \emph{genitive alternation}, the use of \emph{abbreviations} and \emph{lexical relations between words} and \emph{voice change and nominalisation} \cite{Schlegel2020}. Moreover, meaningful \emph{discourse-level connectives} are not represented in MRC benchmarks~\cite{Wu2021}. Similarly, we can evaluate lexical commonsense knowledge, by aligning baseline and intervention instances with \emph{ambiguous prepositional attachments}, \emph{adverbial phrases} or \emph{pronouns} that can only be resolved by understanding the context. In future work, we will curate more challenge sets to evaluate these phenomena. 

\section*{Limitations}

The design of the study limits our findings by design---by relying on lexical and syntactic alterations, we explicitly focus on a narrow set of capabilities, which in fact can be solved by following clear and direct rules. In ``real world'' application scenarios, this narrow subset of capabilities are insufficient to represent the challenges encountered. As such, our study draws fundamental conclusions about the capabilities of transformer-based language models rather than to making specific recommendations which are immediately relevant to practitioners.
Our study also suffers from the inductive dilemma. While we find that our investigated language models follow the trends reported in this paper, due to the empirical nature of this research, this finding is of course not a guarantee that some other neural architecture (transformer-based or otherwise) could still perform well, when tested in our out-of-distribution evaluation settings.

\bibliographystyle{acl_natbib}
\bibliography{references}


\appendix
\section{Reproducibility}
The hyper-parameters used for training are
\begin{itemize}
    \item \textbf{Batch size:} we use batch size of 24 for the \texttt{roberta-base} model.
    \item \textbf{Learning Rate:} We set learning rate to $3^{-5}$, as it was reported to work best for the transformer training. 
    \item \textbf{Train Epochs:} We train on \textsc{SQuAD} and its augmentations for 2 training epochs.
    \item \textbf{Maximal answer length}: we set \verb|max_answer_length=30| when obtaining predictions on the original datasets and \verb|max_answer_length=10| for predictions on the challenge set as they are shorter.
\end{itemize}
The code, environment, a sample of the data, and commands to reproduce the experiments are provided in a separate zip file alongside the submission. All artefacts will be released upon acceptance.
All experiments were carried out on a single NVidia V100 GPU.

\end{document}